\documentclass[conference,10pt]{IEEEtran}

\usepackage[english]{babel}

\usepackage[latin1]{inputenc}

\usepackage{amssymb}

\usepackage{amsmath}

\usepackage{cite} % Loading the cite package will result in citation numbers being automatically sorted and properly "ranged". i.e., [1], [9], [2], [7], [5], [6] (without using cite.sty) will become: [1], [2], [5]--[7], [9] (using cite.sty)
\usepackage{stfloats}  % Gives LaTeX2e the ability to do double column
\usepackage{tikz}
\usepackage{pgf}

\newcommand{\Conf}{\mathrm{Conf}}
\newcommand{\Con}{\mathrm{Con}}
\newcommand{\Cont}{\mathrm{Contr}}
\newcommand{\bel}{\mathrm{bel}}
\newcommand{\pl}{\mathrm{pl}}
\newcommand{\q}{\mathrm{q}}
\newcommand{\betP}{\mathrm{betP}}

\newcommand{\dis}{\mathrm{Dis}}

\newcommand{\Y}{\mathrm{Y}}

\newcommand{\PCR}{\mathrm{PCR}}

\newcommand{\conj}{\mathrm{c}}
\newcommand{\DP}{\mathrm{DP}}
\newcommand{\DS}{\mathrm{DS}}
\newcommand{\NS}{\mathrm{NS}}

\newcommand{\argmax}{\operatornamewithlimits{arg\,max}}

\begin{document}

% paper title
\title{Contradiction measures and specificity degrees of basic belief assignments}
% author names and affiliations
% use a multiple column layout for up to three different
% affiliations

\author{\authorblockN{Florentin Smarandache}
\authorblockA{Math. \& Sciences Dept.\\
University of New Mexico,\\
200 College Road,\\
Gallup, NM 87301, U.S.A.\\
Email: smarand@unm.edu}
\and
\authorblockN{Arnaud Martin}
\authorblockA{IRISA\\
University of Rennes 1\\
Rue \'Edouard Branly BP 30219\\
22302 Lannion, France\\
Email: Arnaud.Martin@univ-rennes1.fr}
\and
\authorblockN{Christophe Osswald}
\authorblockA{E3I2\\
ENSTA Bretagne\\
2, rue Fran{\c c}ois Verny\\
29806 Brest, Cedex 9, France\\
Email: Christophe.Osswald@ensta-bretagne.fr}}

% make the title area
\maketitle

\selectlanguage{english}

\begin{abstract}
In the theory of belief functions, many measures of uncertainty have been introduced. However, it is not always easy to understand what these measures really try to represent. In this paper, we re-interpret some measures of uncertainty in the theory of belief functions. We present some interests and drawbacks of the existing measures. On these observations, we introduce a measure of contradiction. Therefore, we present some degrees of non-specificity and Bayesianity of a mass. We propose a degree of specificity based on the distance between a mass and its most specific associated mass. We also show how to use the degree of specificity to measure the specificity of a fusion rule. Illustrations on simple examples are given.
\end{abstract}


% For peer review papers, you can put extra information on the cover
% page as needed:
% \begin{center} \bfseries EDICS Category: 3-BBND \end{center}
%
% for peerreview papers, inserts a page break and creates the second title.
% Will be ignored for other modes.
\IEEEpeerreviewmaketitle
%==================================================================
\section{Introduction}
\label{sec:Introduction}

The theory of belief functions was first introduced by \cite{Dempster67} in order to represent some imprecise probabilities with {\em upper} and {\em lower probabilities}. Then \cite{Shafer76} proposed a mathematical theory of evidence. \\

Let $\Theta$ be a frame of discernment. A \textit{basic belief assignment} (bba) $m$ is the mapping from elements of the powerset $2^\Theta$ onto $[0,1]$ such that:
\begin{equation}
%\label{close}
\sum_{X\in 2^\Theta} m(X)=1.
\label{normDST}
\end{equation}
The axiom $m(\emptyset)=0$ is often used, but not mandatory. A {\em focal element} $X$ is an element of $2^\Theta$ such that $m(X)\neq 0$.. The difference of a bba with a probability is the domain of definition. A bba is defined on the powerset $2^\Theta$ and not only on $\Theta$. In the powerset, each element is not equivalent in terms of precision. Indeed, $\theta_1 \in \Theta$ is more precise than $\theta_1 \cup \theta_2 \in 2^\Theta$. 

In the case of the DSmT introduced in \cite{Smarandache04}, the bba are defined on an extension of the powerset: the hyper powerset noted $D^\Theta$, formed by the closure of $\Theta$ by union and intersection. The problem of signification of each focal element is the same as in $2^\Theta$. For instance, $\theta_1 \in \Theta$ is less precise than $\theta_1 \cap \theta_2 \in D^\Theta$. In the rest of the paper, we will note $G^\Theta$ for either $2^\Theta$ or $D^\Theta$.

In order to try to quantify the measure of uncertainty such as in the set theory \cite{Hartley28} or in the theory of probabilities \cite{Shannon48}, some measures have been proposed and discussed in the theory of belief functions \cite{Yager83a,Dubois85, Klir94,George96}. However, the domain of definition of the bba does not allow an ideal definition of measure of uncertainty. Moreover, behind the term of uncertainty, different notions are hidden. 

In the section \ref{measureUncertainty}, we present different kinds of measures of uncertainty given in the state of art, we discuss them and give our definitions of some terms concerning the uncertainty. In section \ref{contradiction}, we introduce a measure of contradiction and discuss it. We introduce simple degrees of uncertainty in the section \ref{degreesUncer}, and propose a degree of specificity in the section \ref{degreeSpecificity}. We show how this degree of specificity can be used to measure the specificity of a combination rule.

\section{Measures of uncertainty on belief functions}
\label{measureUncertainty}
In the framework of the belief functions, several functions (we call them {\em belief functions}) are in one to one correspondence with the bba: $\bel$, $\pl$ and $\q$. From these belief functions, we can define several measures of uncertainty. Klir in \cite{Klir94} distinguishes two kinds of uncertainty: the non-specificity and the discord. Hence, we recall hereafter the main belief functions, and some non-specificity and discord measures.

\subsection{Belief functions}
 Hence, the credibility and plausibility functions represent respectively a minimal and maximal belief. The {\em credibility} function is given from a bba for all  $X \in G^\Theta$ by:
\begin{eqnarray}
\bel(X)=\sum_{Y \subseteq X, Y \not\equiv \emptyset} m(Y).
\end{eqnarray}
The {\em plausibility} is given from a bba for all $X \in G^\Theta$ by:
\begin{eqnarray}
\pl(X)=\displaystyle \sum_{Y \in G^\Theta, Y\cap X \not\equiv \emptyset} m(Y).
\end{eqnarray}
The {\em commonality} function is also another belief function given by:
\begin{eqnarray}
\q(X)=\displaystyle \sum_{Y \in G^\Theta, Y\supseteq X} m(Y).
\end{eqnarray}
These functions allow an implicit model of imprecise and uncertain data. However, these functions are monotonic by inclusion: $\bel$ and $\pl$ are increasing, and $\q$ is decreasing. This is the reason why the most of time we use a probability to take a decision. The most used projection into probability subspace is the pignistic probability transformation introduced by \cite{Smets90} and given by:
\begin{eqnarray}
\label{betp}
\betP(X)=\sum_{Y \in G^\Theta, Y \not\equiv \emptyset} \frac{|X \cap Y|}{|Y|} m(Y),
\end{eqnarray}
where $|X|$ is the cardinality of $X$, in the case of the DSmT that is the number of disjoint elements corresponding in the Venn diagram. 

From this probability, we can use the measure of uncertainty given in the theory of probabilities such as the Shannon entropy \cite{Shannon48}, but we loose the interest of the belief functions and the information given on the subsets of the discernment space $\Theta$.

\subsection{Non-specificity}

The non-specificity in the classical set theory is the imprecision of the sets. Such as in \cite{Ristic06}, we define in the theory of belief functions, the non-specificity related to vagueness and non-specificity.

\textbf{Definition} \textit{ The {\em non-specificity} in the theory of belief functions quantifies how a bba $m$ is imprecise.}

The non-specificity of a subset $X$ is defined by Hartley \cite{Hartley28} by $\log_2(|X|)$. This measure was generalized by \cite{Dubois85} in the theory of belief functions by:
\begin{eqnarray}
\label{nonspeDubois}
\NS(m)=\sum_{X \in G^\Theta,~X\not\equiv\emptyset} m(X)\log_2(|X|).
\end{eqnarray}
That is a weighted sum of the non-specificity, and the weights are given by the basic belief in $X$. Ramer in \cite{Ramer87} has shown that it is the unique possible measure of non-specificity in the theory of belief functions under some assumptions such as symmetry, additivity, sub-additivity, continuity, branching and normalization. 

If the measure of the non-specificity on a bba is low, we can consider the bba is specific. Yager in \cite{Yager83a} defined a specificity measure such as:
\begin{equation}
\label{speYager}
 S(m)=\sum_{X\in G^\Theta,~X\not\equiv\emptyset} \frac{m(X)}{|X|}.
\end{equation}

Both definitions corresponded to an accumulation of a function of the basic belief assignment on the focal elements. Unlike the classical set theory, we must take into account the bba in order to quantify (to weight) the belief of the imprecise focal elements. The imprecision of a focal element can of course be given by the cardinality of the element. 

First of all, we must be able to compare the non-specificity (or specificity) between several bba's, event if these bba's are not defined on the same discernment space. That is not the case with the equations \eqref{nonspeDubois} and \eqref{speYager}. The non-specificity of the equation \eqref{nonspeDubois} takes its values in $[0,\log_2(|\Theta|)]$. The specificity of the equation \eqref{speYager} can have values in $[\frac{1}{|\Theta|},1]$. We will show how we can easily define a degree of non-specificity in $[0,1]$. We could also define a degree of specificity from the equation (\ref{speYager}), but that is more complicated and we will later show how we can define a specificity degree.

The most non-specific bba's for both equations \eqref{nonspeDubois} and \eqref{speYager} are the total ignorance bba given by the categorical bba $m_\Theta:m(\Theta)=1$. We have $\NS(m)=\log_2(|\Theta|)$ and $S(m)=\frac{1}{|\Theta|}$. This categorical bba is clearly the most non-specific for us. However, the most specific bba's are the Bayesian bba's. The only focal elements of a Bayesian bba are the simple elements of $\Theta$. On these kinds of bba $m$ we have $\NS(m)=0$ and $S(m)=1$. For example, we take the three Bayesian bba's defined on $\Theta=\{\theta_1,\theta_2,\theta_3\}$ by:
\begin{eqnarray}
m_1(\theta_1)= m_1(\theta_2)= m_1(\theta_3)= 1/3,\\
m_2(\theta_1)= m_2(\theta_2)= 1/2, \, m_2(\theta_3)= 0,\\
m_3(\theta_1)=1, \, m_3(\theta_2)=m_3(\theta_3)= 0.
\end{eqnarray}
We obtain the same non-specificity and specificity for these three bba's.

That hurts our intuition; indeed, we intuitively expect that the bba $m_3$ is the most specific and the $m_1$ is the less specific. We will define a degree of specificity according to a most specific bba that we will introduce.

\subsection{Discord}
Different kinds of discord have been defined as extensions for belief functions of the Shannon entropy, given for the probabilities. Some discord measures have been proposed from plausibility, credibility or pignistic probability:
\begin{eqnarray}
E(m)=-\sum_{X \in G^\Theta} m(X)\log_2(\pl(X)),
\end{eqnarray}
\begin{eqnarray}
C(m)=-\sum_{X \in G^\Theta} m(X)\log_2(\bel(X)),
\end{eqnarray}
\begin{eqnarray}
D(m)=-\sum_{X \in G^\Theta} m(X)\log_2(\betP(X)),
\end{eqnarray}
with $E(m)\leq D(m) \leq C(m)$. We can also give the Shanon entropy on the pignistic probability:
\begin{eqnarray}
-\sum_{X \in G^\Theta} \betP(X)\log_2(\betP(X)).
\end{eqnarray}
Other measures have been proposed, \cite{Klir94} has shown that these measures can be resumed by:
\begin{eqnarray}
-\sum_{X \in G^\Theta} m(X)\log_2(1-\Con_m(X)),
\end{eqnarray}
where $\Con$ is called a conflict measure of the bba $m$ on $X$. However, in our point of view that is not a conflict such presented in \cite{Wierman01}, but a contradiction. We give the both following definitions:

\textbf{Definition} \textit{ 
A {\em contradiction} in the theory of belief functions quantifies how a bba $m$ contradicts itself.}
 
\textbf{Definition} (C1) \textit{The {\em conflict} in the theory of belief functions can be defined by the contradiction between 2 or more bba's.}

In order to measure the conflict in the theory of belief functions, it was usual to use the mass $k$ given by the conjunctive combination rule on the empty set. This rule is given by two basic belief assignments $m_1$ and $m_2$ and for all $X \in G^\Theta$ by:
\begin{eqnarray}
\label{conjunctive}
m_\conj(X)=\displaystyle \sum_{A\cap B =X} m_1(A)m_2(B):=(m_1\oplus m_2 )(X).
\end{eqnarray}
$k=m_\conj(\emptyset)$ can also be interpreted as a non-expected solution. 

In \cite{Yager83a}, Yager proposed another conflict measure from the value of $k$ given by $-\log_2(1-k)$. 

However, as observed in \cite{Liu06}, the weight of conflict given by $k$ (and all the functions of $k$) is not a conflict measure between the basic belief assignments. Indeed this value is completely dependant of the conjunctive rule and this rule is non-idempotent - the combination of identical basic belief assignments leads generally to a positive value of $k$. To highlight this behavior, we defined in \cite{Osswald06} the {\it auto-conflict} which quantifies the intrinsic conflict of a bba. The auto-conflict of order $n$ for one expert is given by: 
\begin{eqnarray}
\label{autoconf}
a_n=\displaystyle \left(\mathop{\oplus}_{i=1}^n m\right)(\emptyset).
\end{eqnarray}
The auto-conflict is a kind of measure of the contradiction, but depends on the order. We studied its behavior in \cite{Martin08}. Therefore we need to define a measure of contradiction independent on the order. This measure is presented in the next section \ref{contradiction}.

\section{A contradiction measure}
\label{contradiction}
The definition of the conflict (C1) involves firstly to measure it on the bba's space and secondly that if the opinions of two experts are far from each other, we consider that they are in conflict. That suggests a notion of distance. That is the reason why in \cite{Martin08}, we give a definition of the measure of conflict between experts assertions through a distance between their respective bba's. The conflict measure between $2$ experts is defined by:
\begin{eqnarray}
\label{2conflict_measure}
\Conf(1,2)=d(m_1,m_2).
\end{eqnarray}
We defined the conflict measure between one expert $i$ and the other $M-1$ experts by:
\begin{eqnarray}
\label{conflict_measure1}
\Conf(i,\mathcal{E})=\frac{1}{M-1}\sum_{j=1, i\neq j}^M \Conf(i,j),
\end{eqnarray}
where $\mathcal{E}=\{1,\ldots, M\}$ is the set of experts in conflict with $i$. Another definition is given by:
\begin{eqnarray}
\label{conflict_measure2}
\Conf(i,M)=d(m_i,\overline{m_M}),
\end{eqnarray}
where $\overline{m_M}$ is the bba of the artificial expert representing the combined opinions of all the experts in $\mathcal{E}$ except $i$. 

We use the distance defined in \cite{Jousselme01}, which is for us the most appropriate, but other distances are possible. See \cite{Florea09} for a comparison of distances in the theory of belief functions. This distance is defined for two basic belief assignments $\textbf{m}_1$ and $\textbf{m}_2$ on $G^\Theta$ by:
\begin{eqnarray}
\label{distance}
d(m_1,m_2)=\sqrt{\frac{1}{2} (\textbf{m}_1-\textbf{m}_2)^T\underline{\underline{D}}(\textbf{m}_1-\textbf{m}_2)},
\end{eqnarray}
where $\underline{\underline{D}}$ is an $G^{|\Theta|}\times G^{|\Theta|}$ matrix based on Jaccard distance whose elements are:
\begin{eqnarray}
\label{DMatrix}
D(A,B)=\left\{
\begin{array}{l}
1, \, \mbox{if} \, A= B=\emptyset,\\
\\
\displaystyle \frac{|A\cap B|}{|A\cup B|}, \, \forall A, B \in G^\Theta.\\
\end{array}
\right.
\end{eqnarray}

However, this measure is a \textit{total conflict} measure. In order to define a contradiction measure we keep the same spirit. First, the contradiction of an element $X$ with respect to a bba $m$ is defined as the distance between the bba's $m$ and $m_X$, where $m_X(X)=1$, $X \in G^\Theta$, is the categorical bba:
\begin{equation}
\label{Eq_cont_element}
 \Cont_m(X)=d(m,m_X),
\end{equation}
where the distance can also be the Jousselme distance on the bba's. The contradiction of a bba is then defined as a weighted contradiction of all the elements $X$ of the considered space $G^\Theta$:
\begin{equation}
\label{Eq_contra_bba}
 \Cont_m=c \sum_{X\in G^\Theta}m(X) d(m,m_X),
\end{equation}

where $c$ is a normalized constant which depends on the type of distance used and on the cardinality of the frame of discernment in order to obtain values in $[0,1]$ as shown in the following illustration.

% If we recall the definition of the conflict such as the contradiction between 2, we can propose another definition of the conflict between two bba's by:
% \begin{equation}
% \Conf(1,2)=\Cont_1+\Cont_2-d(m_1,m_2).
% \end{equation}
% However, this definition of the conflict cannot be extended to more bba's without the lost of the positivity.

%AJOUTER COMMENTAIRES...

\subsection{Illustration}

Here the value $c$ in the equation \eqref{Eq_contra_bba} is equal to 2. First we note that for all categorical bbas $m_Y$, the contradiction given by the equation \eqref{Eq_cont_element} gives $\Cont_{m_Y}(Y)=0$ and the contradiction given by the equation \eqref{Eq_contra_bba} brings also $\Cont_{m_Y}=0$. Considering the bba $m_1(\theta_1)=0.5$ and $m_1(\theta_2)=0.5$, we have  $\Cont_{m_1}=1$. That is the maximum of the contradiction, hence the contraction of a bba takes its values in $[0,1]$.

\begin{figure}[!h]
  \begin{center}
    \caption{Bayesian bba's}
    ~\\

    \renewcommand{\arraystretch}{0.9}
    \begin{tikzpicture}
      \node [shape=circle, draw] at (-2.5,0) {$\begin{array}{c}
          \theta_1 \\ \scriptsize 0.5
        \end{array}$};
      \node [shape=circle, draw] at (0,0) {$\begin{array}{c}
          \theta_2 \\ \scriptsize 0.5
        \end{array}$};
      \node [shape=circle, draw] at (2.5,0) {$\begin{array}{c}
          \theta_3 \\ \scriptsize 0
        \end{array}$};
      \node at (-4,0) {$m_1$:};
    \end{tikzpicture} 

    ~\\

    \renewcommand{\arraystretch}{0.9}
    \begin{tikzpicture}
      \node [shape=circle, draw] at (-2.5,0) {$\begin{array}{c}
          \theta_1 \\ \scriptsize 0.6
        \end{array}$};
      \node [shape=circle, draw] at (0,0) {$\begin{array}{c}
          \theta_2 \\ \scriptsize 0.3
        \end{array}$};
      \node [shape=circle, draw] at (2.5,0) {$\begin{array}{c}
          \theta_3 \\ \scriptsize 0.1
        \end{array}$};
      \node at (-4,0) {$m_2$:};
    \end{tikzpicture}
  \end{center}
\end{figure}
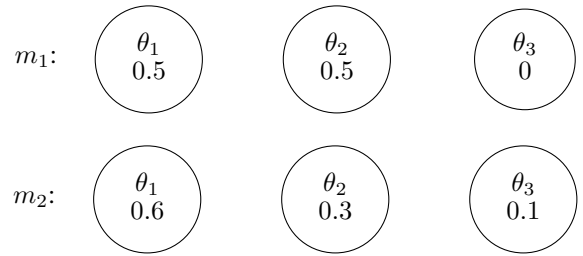

Taking the Bayesian bba given by: $m_2(\theta_1)=0.6$, $m_2(\theta_2)=0.3$, and $m_2(\theta_3)=0.1$. We obtain:
\begin{eqnarray*}
  \Cont_{m_2}(\theta_1)& \simeq&  0.36, \\
  \Cont_{m_2}(\theta_2) & \simeq &0.66, \\ 
  \Cont_{m_2}(\theta_3) &\simeq& 0.79
\end{eqnarray*}

The contradiction for $m_2$ is $\Cont_{m_2}=0.9849$.

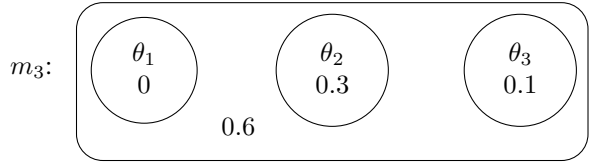
\begin{figure}[!h]
  \begin{center}
    \caption{Non-dogmatic bba}
    ~\\
    \begin{tikzpicture}
      \node [shape=circle, draw] at (-2.5,0) {$\begin{array}{c}
          \theta_1 \\ \scriptsize 0
        \end{array}$};
      \node [shape=circle, draw] at (0,0) {$\mathbf{\begin{array}{c}
          \theta_2 \\ \scriptsize 0.3
        \end{array}}$};
      \node [shape=circle, draw] at (2.5,0) {$\begin{array}{c}
          \theta_3 \\ \scriptsize 0.1
        \end{array}$};
      \draw [rounded corners=10pt] (-3.4,0) -- (-3.4,0.9) -- (3.4,0.9) -- (3.4,-1.2) -- (-3.4,-1.2) -- (-3.4,0);
      \node at (-1.25,-0.75) {$0.6$};
      \node at (-4,0) {$m_3$:};
    \end{tikzpicture}
  \end{center}
\end{figure}

Take $m_3(\{\theta_1,\theta_2,\theta_3\})=0.6$, $m_3(\theta_2)=0.3$, and $m_3(\theta_3)=0.1$; the masses are the same than $m_2$, but the highest one is on $\Theta=\{\theta_1,\theta_2,\theta_3\}$ instead of $\theta_1$. We obtain:
\begin{eqnarray*}
  \Cont_{m_3}(\{\theta_1,\theta_2,\theta_3\}) & \simeq & 0.28, \\
  \Cont_{m_3}(\theta_2)& \simeq & 0.56, \\
  \Cont_{m_3}(\theta_3)&\simeq& 0.71
\end{eqnarray*}
The contradiction for $m_3$ is $\Cont_{m_3}=0.8092$. We can see that the contradiction is lowest thanks to the distance taking into account the imprecision of $\Theta$. 

\begin{figure}[!h]
  \begin{center}
    \caption{Focal elements of cardinality 2}
    ~\\

    \renewcommand{\arraystretch}{0.9}
    \begin{tikzpicture}
      \node at (0:1.5) {$\theta_1$};
      \node at (120:1.5) {$\theta_2$};
      \node at (240:1.5) {$\theta_3$};
      \draw [rounded corners=8pt] (0:2) -- (-15:2) -- (-20:1) -- (0:1);
      \draw [rounded corners=8pt] (120:2) -- (135:2) -- (140:1) -- (120:1);
      \draw (0:2) arc (0:120:2) (120:2);
      \draw (0:1) arc (0:120:1) (120:1);
      \draw [rounded corners=8pt] (120:2.1) -- (105:2.1) -- (100:1.1) -- (120:1.1);
      \draw [rounded corners=8pt] (240:2.1) -- (255:2.1) -- (260:1.1) -- (240:1.1);
      \draw (120:2.1) arc (120:240:2.1) (240:2.1);
      \draw (120:1.1) arc (120:240:1.1) (240:1.1);      
      \draw [rounded corners=8pt] (240:1.9) -- (225:1.9) -- (220:0.9) -- (240:0.9);
      \draw [rounded corners=8pt] (0:1.9) -- (15:1.9) -- (20:0.9) -- (00:0.9);
      \draw (240:1.9) arc (240:360:1.9) (0:1.9);
      \draw (240:0.9) arc (240:360:0.9) (0:0.9);
      \node at (60:1.5) {$0.6$};
      \node at (180:1.5) {$0.1$};
      \node at (300:1.5) {$0.3$};
      \node at (-2.9,0) {$m_4$:};
    \end{tikzpicture}
  \end{center}
\end{figure}
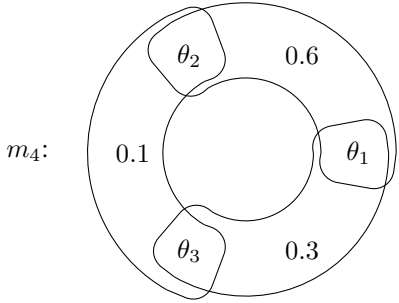

If we consider now the same mass values but on focal elements of cardinality 2:  $m_4(\{\theta_1,\theta_2\})=0.6$, $m_4(\theta_1,\theta_3)=0.3$, and $m_4(\theta_2,\theta_3)=0.1$. We obtain:\\
\begin{eqnarray*}
  \Cont_{m_4}(\{\theta_1,\theta_2\}) & \simeq & 0.29, \\
  \Cont_{m_4}(\{\theta_1,\theta_3\}) & \simeq & 0.53, \\
  \Cont_{m_4}(\{\theta_2,\theta_3\}) & \simeq & 0.65
\end{eqnarray*}
The contradiction for $m_4$ is $\Cont_{m_4}=0.80$. \\

Fewer of focal elements there are, smaller the contradiction of the bba will be, and more the focal elements are precise, higher the contradiction of the bba will be.

\section{Degrees of uncertainty}
\label{degreesUncer}
We have seen in the section \ref{measureUncertainty} that the measure non-specificity given by the equation \eqref{nonspeDubois} take its values in a space dependent on the size of the discernment space  $\Theta$. Indeed, the measure of non-specificity takes its values in $[0,\log_2(|\Theta|)]$. 

In order to compare some non-specificity measures in an absolute space, we define a degree of non-specificity from the equation \eqref{nonspeDubois} by:
\begin{eqnarray}
\label{degrenonspeDubois}
\begin{array}{rcl}
\delta_{\NS}(m)&=&\displaystyle \sum_{X \in G^\Theta,~X\not\equiv\emptyset} m(X)\frac{\log_2(|X|)}{\log_2(|\Theta|)}\\
&=& \displaystyle \sum_{X \in G^\Theta,~X\not\equiv\emptyset} m(X)\log_{|\Theta|}(|X|).
\end{array}
\end{eqnarray}
Therefore, this degree takes its values into $[0,1]$ for all bba's $m$, independently of the size of discernment. We still have $\delta_{\NS}(m_\Theta)=1$, where $m_\Theta$ is the categorical bba giving the total ignorance. Moreover, we obtain $\delta_{\NS}(m)=0$ for all Bayesian bba's.

From the definition of the degree of non-specificity, we can propose a degree of specificity such as:
\begin{eqnarray}
\label{degrebayesianity}
\begin{array}{rcl}
\delta_{B}(m)&=&\displaystyle 1-\sum_{X \in G^\Theta,~X\not\equiv\emptyset} m(X)\frac{\log_2(|X|)}{\log_2(|\Theta|)}\\
&=& \displaystyle 1-\sum_{X \in G^\Theta,~X\not\equiv\emptyset} m(X)\log_{|\Theta|}(|X|).
\end{array}
\end{eqnarray}

As we observe already the degree of non-specificity given by the equation \eqref{degrebayesianity} does not really measure the specificity but the Bayesianity of the considered bba. This degree is equal to 1 for Bayesian bba's and not one for other bba's.

\textbf{Definition} \textit{ The {\em Bayesianity} in the theory of belief functions quantify how far a bba $m$ is from a probability.}

We illustrate these degrees in the next subsection.

\subsection{Illustration}
In order to illustrate and discuss the previous introduced degrees we take some examples given in the table~\ref{table_bayesianity}. The bba's are defined on $2^\Theta$ where $\Theta=\{\theta_1,\theta_2,\theta_3\}$. We only consider here non-Bayesian bba's, else the values are still given hereinbefore.

We can observe for a given sum of basic belief on the singletons of $\Theta$ the Bayesianity degree can change according to the basic belief on the other focal elements. For example, the degrees are the same for $m_2$ and $m_3$, but different for $m_4$. For the bba $m_4$, note that the sum of the basic beliefs on the singletons is equal to the basic belief on the ignorance. In this case the Bayesianity degree is exactly 0.5. That is conform to the intuitive signification of the Bayesianity. If we look $m_5$ and $m_6$, we first note that there is no basic belief on the singletons. As a consequence, the Bayesianity is weaker. Moreover, the bba $m_5$ is naturally more Bayesian than $m_6$ because of the basic belief on $\Theta$.

\begin{table}
% increase table row spacing, adjust to taste
\renewcommand{\arraystretch}{1.3}
\caption{Evaluation of Bayesianity on examples}
\label{table_bayesianity}
\begin{center}
\begin{tabular}{|c|c|c|c|c|c|c|c|}
\hline
 & $m_1$ & $m_2$ & $m_3$ & $m_4$ & $m_5$ & $m_6$ & $m_\Theta$\\
\hline
$\theta_1$ &  0.4& 0.3 & 0.1 & 0.3& 0 & 0 & 0  \\
\hline
$\theta_2$ &  0.1 & 0.1 & 0.3 & 0.1 & 0 & 0 & 0  \\
\hline
$\theta_3$ & 0.1 & 0.1 & 0.1 & 0.1 & 0 & 0 & 0  \\
\hline
$\theta_1 \cup \theta_2$ & 0.3 & 0.3 & 0.5 & 0 & 0.6 & 0.6 & 0 \\
\hline
$\theta_1 \cup \theta_3$  & 0.1 & 0.2 & 0 & 0 & 0.4 & 0 & 0 \\
\hline
$\theta_2 \cup \theta_3$ & 0 & 0 & 0 & 0 & 0 & 0 & 0  \\
\hline
$\Theta$ & 0 & 0 & 0 & 0.5 & 0 & 0.4 & 1  \\
\hline
\hline
$\delta_B$ & 0.75 &  0.68 & 0.68 & 0.5 & 0.37 & 0.23 & 0   \\
\hline
$\delta_{\NS}$ & 0.25 & 0.32 & 0.32 & 0.5 & 0.63 & 0.77 & 1   \\
\hline
\end{tabular}
\end{center}
\end{table}

We must add that these degrees are dependent on the cardinality of the frame of discernment for non Bayesian bba's. Indeed, if we consider the given bba $m_1$ with $\Theta=\{\theta_1,\theta_2,\theta_3\}$, the degree $\delta_B=0.75$. Now if we consider the same bba with $\Theta=\{\theta_1,\theta_2,\theta_3, \theta_4\}$ (no beliefs are given on $\theta_4$), the Bayesianity degree is $\delta_B=0.80$. The larger is the frame, the larger becomes the Bayesianity degree. 

\section{Degree of specificity}
\label{degreeSpecificity}
In the previous section, we showed the importance to consider a degree instead of a measure. Moreover, the measures of specificity and non-specificity given by the equations (\ref{speYager}) and (\ref{nonspeDubois}) give the same values for every Bayesian bba's as seen on the examples of the section \ref{measureUncertainty}. 

We introduce here a degree of specificity based on comparison with the bba the most specific. This degree of specificity is given by:
\begin{equation}
\label{degreeSpe}
 \delta_S(m)=1-d(m,m_s),
\end{equation}
where $m_s$ is the bba the most specific associated to $m$ and $d$ is a distance defined onto $[0,1]$. Here we also choose the Jousselme distance, the most appropriated on the bba's space, with values onto $[0,1]$. This distance is dependent on the size of the space $G^\Theta$, we have to compare the degrees of specificity for bba's defined from the same space. Accordingly, the main problem is to define the bba the most specific associated to $m$.  

\subsection{The most specific bba} 
In the theory of belief functions, several partial orders have been proposed in order to compare the bba's \cite{Dubois86}. These partial ordering are generally based on the comparisons of their plausibilities or their communalities, specially in order to find the least-committed bba. 
However, comparing bba's with plausibilities or communality can be complex and without unique solution. 

The problem to find the most specific bba associated to a bba $m$ does not need to use a partial ordering. We limit the specific bba's to the categorical bba's: $m_X(X)=1$ where $X\in G^\Theta$ and we will use the following axiom and proposition:

\textbf{Axiom} \textit{For two categorical bba's $m_X$ and $m_Y$, $m_X$ is more specific than $m_Y$ if and only if $|X| < |Y|$. }

In case of equality, the both bba's are \textit{isospecific}. 

\textbf{Proposition} \textit{If we consider two isospecific bba's $m_X$ and $m_Y$, the Jousselme distance between every bba $m$ and $m_X$ is equal to the Jousselme distance between $m$ and $m_Y$:
  \begin{equation}
    \forall m,~d(m,m_X)=d(m,m_Y)
  \end{equation}
  if and only if $m(X)=m(Y)$.}

\textbf{Proof} \textit{The proof is obvious considering the equations \eqref{distance} and \eqref{DMatrix}. As the both bba's $m_X$ and $m_Y$ are categoric there is only one non null term in the difference of vectors $m-m_X$ and $m-m_Y$. These both terms $a_X$ and $a_Y$ are equal, because $m_X$ and $m_Y$ are isospecific and so according to the equation \eqref{DMatrix} $D(Z,X)=D(Z,Y)$ $\forall Z \in G^\Theta$. Therefore $m(X)=m(Y)$, that proves the proposition 
}\hfill$\square$

We define \textit{the most specific bba} $m_s$ associated to a bba $m$ as a categorical bba as follows: $m_s(X_{\max})=1$ where $X_{\max}\in G^\Theta$.

Therefore, the matter is now how to find $X_{\max}$. We propose two approaches:

\begin{description}
\item[ {\bf First approach, Bayesian case}] ~\\
\vspace{-3mm}

If $m$ is a Bayesian bba we define $X_{\max}$ such as:
\begin{equation}
\label{bayesianspecific}
 X_{\max}=\argmax(m(X), \, X\in \Theta).
\end{equation}

If there exist many $X_{\max}$ ({\em i.e.} having the same maximal bba and giving many isospecific bba's), we can take any of them. Indeed, according to the previous proposition, the degree of specificity of $m$ will be the same with $m_s$ given by either $X_{\max}$ satisfying (\ref{bayesianspecific}).

\item[{\bf First approach, non-Bayesian case}] ~\\
\vspace{-3mm}

If $m$ is a non-Bayesian bba, we can define $X_{\max}$ in a similar way such as:
\begin{equation}
 X_{\max}=\argmax\left(\frac{m(X)}{|X|},\,X\!\!\in G^\Theta,\,X\!\not\equiv\!\emptyset\right)\!\!.
\end{equation}
In fact, this equation generalizes the equation \eqref{bayesianspecific}. However, in this case we can also have several $X_{\max}$ not giving isospecific bba's. Therefore, we choose one of the more specific bba, {\em i.e.} believing in the element with the smallest cardinality. Note also that we keep the terms of Yager from the equation~\eqref{speYager}.

\item[{\bf Second approach}]   ~\\
\vspace{-3mm}

   Another way in the case of non-Bayesian bba $m$ is to transform $m$ into a Bayesian bba, thanks to one of the probability transformation such as the pignistic probability. Afterwards, we can apply the previous Bayesian case. With this approach, the most specific bba associated to a bba $m$ is always a categorical bba such as: $m_s(X_{\max})=1$ where $X_{\max}\in \Theta$ and not in $G^\Theta$.
\end{description}

\subsection{Illustration}

In order to illustrate this degree of specificity we give some examples. The table \ref{table_bayesianSpeDegree} gives the degree of specificity for some Bayesian bba's. The smallest degree of specificity of a Bayesian bba is obtained for the uniform distribution ($m_1$), and the largest degree of specificity is of course obtain for categorical bba ($m_8$). 

The degree of specificity increases when the differences between the mass of the largest singleton and the masses of other singletons are getting bigger: $\delta_S(m_3)<\delta_S(m_4)<\delta_S(m_5)<\delta_S(m_6)$. In the case when one has three disjoint singletons and the largest mass of one of them is 0.45 (on $\theta_1$), the minimum degree of specificity is reached when the masses of $\theta_2$ and $\theta_3$ are getting further from the mass of $\theta_1$ ($m_6$). If two singletons have the same maximal mass, bigger this mass is and bigger is the degree of specificity: $\delta_S(m_2)<\delta_S(m_3)$.

\begin{table}
% increase table row spacing, adjust to taste
\renewcommand{\arraystretch}{1.3}
\caption{Illustration of the degree of specificity on Bayesian bba.}
\label{table_bayesianSpeDegree}
\begin{center}
\begin{tabular}{|c|c|c|c|c|}
\hline
 & $\theta_1$ & $\theta_2$ & $\theta_3$  &   $\delta_S$\\
\hline
$m_1$ &  1/3 & 1/3 & 1/3 & 0.423  \\
\hline
$m_2$ & 0.4 & 0.4 & 0.2 & 0.471  \\
\hline
$m_3$ & 0.45 & 0.45 & 0.10 & 0.493  \\
\hline
$m_4$ & 0.45 & 0.40 & 0.15 & 0.508  \\
\hline
$m_5$ & 0.45 & 0.3 & 0.25 & 0.523  \\
\hline
$m_6$ & 0.45 & 0.275 & 0.275 & 0.524  \\
\hline
$m_7$ & 0.6 & 0.3 & 0.1 & 0.639  \\
\hline
 $m_8$ &  1& 0 & 0 & 1 \\
\hline
\end{tabular}
\end{center}
\end{table}

In the case of non-Bayesian bba, we first take a simple example:
\begin{eqnarray}
 m_1(\theta_1)=0.6,&\!\!\!\!& m_1(\theta_1 \cup \theta_2)=0.4 \\
 m_2(\theta_1)=0.5,&\!\!\!\!& m_2(\theta_1 \cup \theta_2)=0.5. 
\end{eqnarray}
For these two bba's $m_1$ and $m_2$, both proposed approaches give the same most specific bba: $m_{\theta_1}$. We obtain $\delta_S(m_1)=0.7172$ and $\delta_S(m_2)=0.6465$. Therefore, $m_1$ is more specific than $m_2$. Remark that these degrees are the same if we consider the bba's defined on $2^{\Theta_2}$ and $2^{\Theta_3}$, with $\Theta_2=\{\theta_1,\theta_2\}$ and $\Theta_3=\{\theta_1,\theta_2,\theta_3\}$. If we now consider Bayesian bba $m_3(\theta_1)=m_3(\theta_2)=0.5$, the associated degree of specificity is $\delta_S(m_3)=0.5$. As expected by intuition, $m_2$ is more specific than $m_3$.

We consider now the following bba:
\begin{eqnarray}
 m_4(\theta_1)=0.6, \quad m_1(\theta_1 \cup \theta_2 \cup \theta_3)=0.4.
\end{eqnarray}

The most specific bba is also $m_{\theta_1}$, and we have $\delta_S(m_4)=0.6734$. This degree of specificity is naturally smaller than $\delta_S(m_1)$ because of the mass 0.4 on a more imprecise focal element.

Let's now consider the following example:
\begin{equation}
 m_5(\theta_1 \cup \theta_2)= 0.7, \quad  m_5(\theta_1 \cup \theta_3)= 0.3.
\end{equation}
We do not obtain the same most specific bba with both proposed approaches: the first one will give the categorical bba $m_{\theta_1 \cup \theta_2}$ and the second one $m_{\theta_1}$. Hence, the first degree of specificity is $\delta_S(m_5)=0.755$ and the second one is $\delta_S(m_5)=0.111$. We note that the second approach produces naturally some smaller degrees of specificity.

\subsection{Application to measure the specificity of a combination rule}
We propose in this section to use the proposed degree of specificity in order to measure the quality of the result of a combination rule in the theory of belief functions. Indeed, many combination rules have been developed to merge the bba's \cite{Martin07,Smets07}. The choice of one of them is not always obvious. For a special application, we can compare the produced results of several rules, or try to choose according to the special proprieties wanted for an application. We also proposed to study the comportment of the rules on generated bba's \cite{Osswald06}. However, no real measures have been used to evaluate the combination rules. Hereafter, we only show how we can use the degree of specificity to evaluate and compare the combination rules in the theory of belief functions. A complete study could then be done for example on generated bba's. We recall here the used combination rules, see \cite{Martin07} for their description.

The {\em Dempster rule} is the normalized conjunctive combination rule of the equation \eqref{conjunctive} given for two basic belief assignments $m_1$ and $m_2$ and for all $X \in G^\Theta$, $X\not\equiv\emptyset$ by:
\begin{eqnarray}
m_\DS(X)=\displaystyle \frac{1}{1-k}\sum_{A\cap B =X} m_1(A)m_2(B).
\end{eqnarray}
where $k$ is either $m_\conj(\emptyset)$ or the sum of the masses of the elements of $\emptyset$ equivalence class for $D^\Theta$.

The {\em Yager rule} transfers the global conflict on the total ignorance $\Theta$:
\begin{equation}
m_\Y(X) = \left\{\begin{array}{ll}
  m_\conj(X) & \mbox{if~} X \in 2^\Theta \setminus \{\emptyset, \Theta\} \\
  m_\conj(\Theta)+m_\conj(\emptyset) &  \mbox{if~} X = \Theta \\
  0  &  \mbox{if~} X = \emptyset
\end{array}\right.
\end{equation}

The disjunctive combination rule is given for two basic belief assignments $m_1$ and $m_2$ and for all $X \in G^\Theta$ by:
\begin{eqnarray}
m_\dis(X)=\displaystyle \sum_{A\cup B =X} m_1(A)m_2(B).
\end{eqnarray}
The Dubois and Prade rule is given for two basic belief assignments $m_1$ and $m_2$ and for all $X \in G^\Theta$, $X\not\equiv\emptyset$ by:
\begin{equation}
\label{DP}
m_\DP(X)= \!\!\!\!\sum_{A \cap B = X}\!\!\! m_1(A) m_2(B) + \!\!\!\!\!\sum_{
\begin{array}{c}
\scriptstyle A \cup B=X\\
\scriptstyle A \cap B \equiv \emptyset
\end{array}} \!\!\!\!\! m_1(A)m_2(B).
\end{equation}
The $\PCR$ rule is given for two basic belief assignments $m_1$ and $m_2$ and for all $X \in G^\Theta$, $X\not\equiv \emptyset$ by:
\begin{eqnarray}
\label{DSmTcombination}
\begin{array}{l}
m_\PCR(X)=m_\conj(X)~+\\
\displaystyle
\sum_{\begin{array}{l}
\scriptstyle Y\in G^\Theta, \\
\scriptstyle X\cap Y \equiv \emptyset 
\end{array}} \!\!\!\!\!\left(\frac{m_1(X)^2 m_2(Y)}{m_1(X) \!+\!
  m_2(Y)}+\frac{m_2(X)^2 m_1(Y)}{m_2(X) \!+\! m_1(Y)}\!\right)\!\!,
\end{array}
\end{eqnarray}

The principle is very simple: compute the degree of specificity of the bba's you want combine, then calculate the degree of specificity obtained on the bba after the chosen combination rule. The degree of specificity can be compared to the degrees of specificity of the combined bba's. 

In the following example given in the table \ref{table_fusionBayes} we combine two Bayesian bba's with the discernment frame $\Theta=\{\theta_1,\theta_2,\theta_3\}$. Both bba's are very contradictory. The values are rounded up. The first approach gives the same degree of specificity than the second one except for the rules $m_\dis$, $m_\DP$ and $m_\Y$. We observe that the smallest degree of specificity is obtained for the conjunctive rule because of the accumulated mass on the empty set not considered in the calculus of the degree. The highest degree of specificity is reached for the Yager rule, for the same reason. That is the only rule given a degree of specificity superior to $\delta_S(m_1)$ and to $\delta_S(m_2)$. The second approach shows well the loss of specificity with the rules $m_\dis$, $m_\Y$ and $m_\DP$ having a cautious comportment. With the example, the degree of specificity obtained by the combination rules are almost all inferior to $\delta_S(m_1)$ and to $\delta_S(m_2)$, because the bba's are very conflicting. If the degrees of specificity of the rule such as $m_\DS$ and $m_\PCR$ are superior to $\delta_S(m_1)$ and to $\delta_S(m_2)$, that means that the bba's are not in conflict.

\begin{table}
% increase table row spacing, adjust to taste
\renewcommand{\arraystretch}{1.3}
\caption{Degrees of specificity for combination rules on Bayesian bba's.}
\label{table_fusionBayes}
\begin{center}
\hspace{-5mm}\begin{tabular}{|c|c|c|c|c|c|c|c|c|}
\hline
 & $m_1$ & $m_2$ & $m_\conj$ & $m_\DS$ & $m_\Y$ & $m_\dis$ & $m_\DP$ & \!\!$m_\PCR$\!\!\\
\hline
$\emptyset$ &  0 & 0 & 0.76 & 0 & 0 & 0 & 0 & 0  \\
\hline
$\theta_1$ &  0.6& 0.2 & 0.12 & 0.50 & 0.12 & 0.12 & 0.12 & 0.43  \\
\hline
$\theta_2$ &  0.1 & 0.6 & 0.06 & 0.25 & 0.06 & 0.06 & 0.06 & 0.37  \\
\hline
$\theta_3$ & 0.3 & 0.2 & 0.06 &0.25 & 0.06 & 0.06 & 0.06  & 0.20\\
\hline
\!\!\!\!\!\! $\theta_1 \cup \theta_2$ \!\!\!\!\!\!  & 0 & 0 & 0 & 0 & 0 & 0.38 & 0.38& 0  \\
\hline
\!\!\!\!\!\! $\theta_1 \cup \theta_3$ \!\!\!\!\!\!  & 0 & 0 & 0 & 0 & 0 & 0.18 & 0.18& 0  \\
\hline
\!\!\!\!\!\! $\theta_2 \cup \theta_3$ \!\!\!\!\!\!  & 0 & 0 & 0 & 0 & 0 & 0.20 & 0.20 & 0  \\
\hline
$\Theta$ & 0 & 0 & 0 & 0 & 0.76 & 0 & 0  & 0 \\
\hline
\hline
\!\!\!\!$m_s$ 1- \!\!\!\!& $m_{\theta_1}$ &  $m_{\theta_2}$ & $m_{\theta_1}$ & $m_{\theta_1}$ & \!\!\!\!\!\!$m_{\Theta}$\!\!\!\!\!\! & $m_{\theta_1 \cup \theta_2}$ & \!\!\!\!\!\!$m_{\theta_1 \cup \theta_2}$\!\!\!\!\!\!&  $m_{\theta_1}$  \\
\hline
\!\!\!\!$m_s$ 2- \!\!\!\!& $m_{\theta_1}$ &  $m_{\theta_2}$ & $m_{\theta_1}$ & $m_{\theta_1}$ & \!\!\!\!\!\!$m_{\theta_1}$\!\!\!\!\!\! & $m_{\theta_1}$ & \!\!\!\!\!\!$m_{\theta_1}$\!\!\!\!\!\!&  $m_{\theta_1}$  \\
\hline
\!\!\!\!$\delta_S$ 1- \!\!\!\!& \!\!\!\!\!0.639 \!\!\!\!\!&\!\!\!\!\! 0.655\!\!\!\!\! & \!\!\!\!\!0.176 \!\!\!\!\!& \!\!\!\!\!0.567 \!\!\!\!\!& \!\!\!\!\!0.857 \!\!\!\!\!& \!\!\!\!\!0.619 \!\!\!\!\!& \!\!\!\!\!0.619\!\!\!\!\! & \!\!\!\!\!0.497 \!\!\!\!\!  \\
\hline
\!\!\!\!$\delta_S$ 2- \!\!\!\!& \!\!\!\!\!0.639 \!\!\!\!\!&\!\!\!\!\!\!\! 0.655\!\!\!\!\!\!\! & \!\!\!\!\!\!\!0.176 \!\!\!\!\!\!\!& \!\!\!\!\!\!\!0.567 \!\!\!\!\!\!\!& \!\!\!\!\!0.457 \!\!\!\!\!& \!\!\!\!\!\!\!0.478 \!\!\!\!\!\!\!& \!\!\!\!\!0.478\!\!\!\!\! & \!\!\!\!\!0.497 \!\!\!\!\!  \\
\hline
\end{tabular}
\end{center}
\end{table}

Let's consider now a simple non-Bayesian example in table~\ref{table_fusionNonBayes}.

\begin{figure}[!h]
  \begin{center}
    \caption{Two non-Bayesian bba's}
    ~\\

    \renewcommand{\arraystretch}{0.9}
    \begin{tikzpicture}
      \node [shape=circle, draw, line width = 2pt] at (-2.5,0) {$\begin{array}{c}
          \theta_1 \\ \scriptsize 0.4
        \end{array}$};
      \node [shape=circle, draw] at (0,0) {$\begin{array}{c}
          \theta_2 \\ \scriptsize 0.1
        \end{array}$};
      \node [shape=circle, draw] at (2.5,0) {$\begin{array}{c}
          \theta_3 \\ \scriptsize 0.3
        \end{array}$};
      \draw[rounded corners=10pt] (-3.5,0) -- (-3.5,1) -- (1,1) -- (1,-1) -- (-3.5,-1) -- (-3.5,0);
      \node at (-1.25,0) {$0.2$};
      \node at (-4,0) {$m_1$:};
    \end{tikzpicture}

    ~\\

    \begin{tikzpicture}
      \node [shape=circle, draw] at (-2.5,0) {$\begin{array}{c}
          \theta_1 \\ \scriptsize 0.2
        \end{array}$};
      \node [shape=circle, draw, line width = 2pt] at (0,0) {$\mathbf{\begin{array}{c}
          \theta_2 \\ \scriptsize 0.3
        \end{array}}$};
      \node [shape=circle, draw] at (2.5,0) {$\begin{array}{c}
          \theta_3 \\ \scriptsize 0.1
        \end{array}$};
      \draw[rounded corners=10pt] (-3.3,0) -- (-3.3,1) -- (0.8,1) -- (0.8,-1) -- (-3.3,-1) -- (-3.3,0);
      \node at (-1.3,0) {$0.1$};
      \draw [rounded corners=10pt] (3.3,0) -- (3.3,0.9) -- (-0.9,0.9) -- (-0.9,-0.9) -- (3.3,-0.9) -- (3.3,0);
      \node at (1.3,0) {$0.2$};
      \draw [rounded corners=10pt] (-3.5,0) -- (-3.5,1.1) -- (3.5,1.1) -- (3.5,-1.5) -- (-3.5,-1.5) -- (-3.5,0);
      \node at (0,-1.2) {$0.1$};
      \node at (-4,0) {$m_2$:};
    \end{tikzpicture}
  \end{center}
\end{figure}
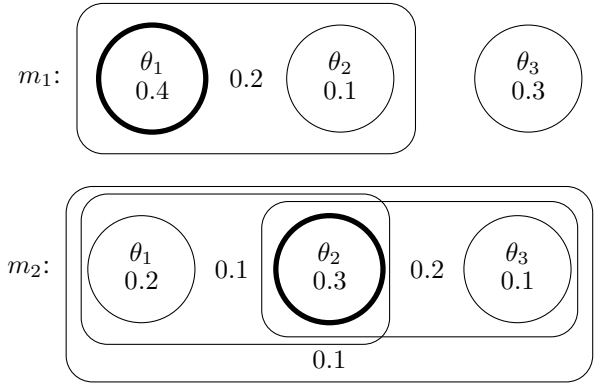

\begin{table}
% increase table row spacing, adjust to taste
\renewcommand{\arraystretch}{1.3}
\caption{Degrees of specificity for combination rules on non-Bayesian bba's.}
\label{table_fusionNonBayes}
\begin{center}
\begin{tabular}{|c|c|c|c|c|c|c|c|c|}
\hline
 & $m_1$ & $m_2$ & $m_\conj$ & $m_\DS$ & $m_\Y$ & $m_\dis$ & $m_\DP$ & \!\!$m_\PCR$\!\!\\
\hline
$\emptyset$ &  0 & 0 & 0.47 & 0 & 0 & 0 & 0 & 0  \\
\hline
$\theta_1$ &  0.4& 0.2 & 0.2 & 0.377 & 0.2 & 0.08 & 0.2 & 0.39  \\
\hline
$\theta_2$ &  0.1 & 0.3 & 0.17 & 0.321 & 0.17 & 0.03 & 0.17 & 0.28  \\
\hline
$\theta_3$ & 0.3 & 0.1 & 0.12 &0.226 & 0.12 & 0.03 & 0.12  & 0.24\\
\hline
\!\!\!\!\!\! $\theta_1 \cup \theta_2$ \!\!\!\!\!\!  & 0.2 & 0.1 & 0.04 & 0.076 & 0.04 & 0.31 & 0.18& 0.06  \\
\hline
\!\!\!\!\!\! $\theta_1 \cup \theta_3$ \!\!\!\!\!\!  & 0 & 0 & 0 & 0 & 0 & 0.1 & 0.1& 0  \\
\hline
\!\!\!\!\!\! $\theta_2 \cup \theta_3$ \!\!\!\!\!\!  & 0 & 0.2 & 0 & 0 & 0 & 0.18 & 0.1 & 0.03  \\
\hline
$\Theta$ & 0 & 0.1 & 0 & 0 & 0.47 & 0.27 & 0.13  & 0 \\
\hline
\hline
\!\!\!\!$m_s$ 1- \!\!\!\!& $m_{\theta_1}$ &  $m_{\theta_2}$ & $m_{\theta_1}$ & $m_{\theta_1}$ & \!\!\!\!\!\!$m_{\theta_1}$\!\!\!\!\!\! & $m_{\theta_1 \cup \theta_2}$ & \!\!\!\!\!\!$m_{\theta_1}$\!\!\!\!\!\!&  $m_{\theta_1}$  \\
\hline
\!\!\!\!$m_s$ 2- \!\!\!\!& $m_{\theta_1}$ &  $m_{\theta_2}$ & $m_{\theta_1}$ & $m_{\theta_1}$ & \!\!\!\!\!\!$m_{\theta_1}$\!\!\!\!\!\! & $m_{\theta_1}$ & \!\!\!\!\!\!$m_{\theta_1}$\!\!\!\!\!\!&  $m_{\theta_1}$  \\
\hline
\!\!\!\!$\delta_S$ 1- \!\!\!\!& \!\!\!\!\!0.553 \!\!\!\!\!&\!\!\!\!\! 0.522\!\!\!\!\! & \!\!\!\!\!0.336 \!\!\!\!\!& \!\!\!\!\!0.488 \!\!\!\!\!& \!\!\!\!\!0.389 \!\!\!\!\!& \!\!\!\!\!0.609 \!\!\!\!\!& \!\!\!\!\!0.428\!\!\!\!\! & \!\!\!\!\!0.497 \!\!\!\!\!  \\
\hline
\!\!\!\!$\delta_S$ 2- \!\!\!\!& \!\!\!\!\!0.553 \!\!\!\!\!&\!\!\!\!\!\!\! 0.522\!\!\!\!\!\!\! & \!\!\!\!\!\!\!0.336 \!\!\!\!\!\!\!& \!\!\!\!\!\!\!0.488 \!\!\!\!\!\!\!& \!\!\!\!\!0.389 \!\!\!\!\!& \!\!\!\!\!\!\!0.456 \!\!\!\!\!\!\!& \!\!\!\!\!0.428\!\!\!\!\! & \!\!\!\!\!0.497 \!\!\!\!\!  \\
\hline
\end{tabular}
\end{center}
\end{table}

\section{Conclusion}
First, we propose in this article a reflection on the measures on uncertainty in the theory of belief functions. A lot of measures have been proposed to quantify different kind of uncertainty such as the specificity - very linked to the imprecision - and the discord. The discord, we do not have to confuse with the conflict, is for us a contradiction of a source (giving information with a bba in the theory of belief functions) with oneself. We distinguish the contradiction and the conflict that is the contradiction between 2 or more bba's. We introduce a measure of contradiction for a bba based on the weighted average of the conflict between the bba and the categorical bba's of the focal elements.

The previous proposed specificity or non-specificity measures are not defined on the same space. Therefore that is difficult to compare them. That is the reason why we propose the use of degree of uncertainty. Moreover these measures give some counter-intuitive results on Bayesian bba's. We propose a degree of specificity based on the distance between a mass and its most specific associated mass that we introduce. This most specific associated mass can be obtained by two ways and give the nearest categorical bba for a given bba. We propose also to use the degree of specificity in order to measure the specificity of a fusion rule. That is a tool to compare and evaluate the several combination rules given in the theory of belief functions.

\subsection*{Acknowledgments}

The authors want to thanks {\em \sc Brest Metropole Oc\'eane} and {\em
  \sc ENSTA Bretagne} for funding this collaboration and providing
them an excellent research environment during spring 2010.

% An example of a double column floating figure using two subfigures.
%(The subfigure.sty package must be loaded for this to work.)
% The subfigure \label commands are set within each subfigure command, the
% \label for the overall fgure must come after \caption.
% \hfil must be used as a separator to get equal spacing
%
%\begin{figure*}
%\centerline{\subfigure[Case I]{\includegraphics[width=2.5in]{subfigcase1}
% where an .eps filename suffix will be assumed under latex, 
% and a .pdf suffix will be assumed for pdflatex
%\label{fig_first_case}}
%\hfil
%\subfigure[Case II]{\includegraphics[width=2.5in]{subfigcase2}
% where an .eps filename suffix will be assumed under latex, 
% and a .pdf suffix will be assumed for pdflatex
%\label{fig_second_case}}}
%\caption{Simulation results}
%\label{fig_sim}
%\end{figure*}

% An example of a floating table. Note that, for IEEE style tables, the 
% \caption command should come BEFORE the table. Table text will default to
% \footnotesize as IEEE normally uses this smaller font for tables.
% The \label must come after \caption as always.
%
%\begin{table}
%% increase table row spacing, adjust to taste
%\renewcommand{\arraystretch}{1.3}
%\caption{An Example of a Table}
%\label{table_example}
%\begin{center}
%% Some packages, such as MDW tools, offer better commands for making tables
%% than the plain LaTeX2e tabular which is used here.
%\begin{tabular}{|c||c|}
%\hline
%One & Two\\
%\hline
%Three & Four\\
%\hline
%\end{tabular}
%\end{center}
%\end{table}

% use section* for acknowledgement
% optional entry into table of contents (if used)
%\addcontentsline{toc}{section}{Acknowledgment}

\bibliographystyle{IEEEtran}
% argument is your BibTeX string definitions and bibliography database(s)

% that's all folks
\end{document}